\documentclass[11pt,a4paper]{article}
\usepackage[hyperref]{emnlp2020}

\usepackage{graphicx}
\usepackage{subcaption}

\usepackage{times}
\usepackage{latexsym}

\usepackage{microtype}
\usepackage{booktabs}
\usepackage{changepage}

\aclfinalcopy 
\def\aclpaperid{111} 

\usepackage{adjustbox}

\usepackage{amsthm}
\theoremstyle{definition}

\newcommand\norm[1]{\left\lVert#1\right\rVert}

\usepackage{algorithm}
\usepackage{algorithmic}

\usepackage{hyperref}

\expandafter\def\expandafter\UrlBreaks\expandafter{\UrlBreaks
  \do\a\do\b\do\c\do\d\do\e\do\f\do\g\do\h\do\i\do\j%
  \do\k\do\l\do\m\do\n\do\o\do\p\do\q\do\r\do\s\do\t%
  \do\u\do\v\do\w\do\x\do\y\do\z\do\A\do\B\do\C\do\D%
  \do\E\do\F\do\G\do\H\do\I\do\J\do\K\do\L\do\M\do\N%
  \do\O\do\P\do\Q\do\R\do\S\do\T\do\U\do\V\do\W\do\X%
  \do\Y\do\Z}

\usepackage{amsmath,amsfonts,bm}

\def\eqref#1{equation~\ref{#1}}

\def\1{\bm{1}}

\DeclareMathAlphabet{\mathsfit}{\encodingdefault}{\sfdefault}{m}{sl}
\SetMathAlphabet{\mathsfit}{bold}{\encodingdefault}{\sfdefault}{bx}{n}

\newcommand{\E}{\mathbb{E}}

\newcommand{\R}{\mathbb{R}}

\title{Aging Memories Generate More Fluent Dialogue Responses\\with Memory Augmented Neural Networks}
\author{Omar U. Florez \\
  Conversational AI Research \\
  Capital One \\\And
  Erick Mueller \\
  Conversational AI Research \\
  Capital One \\}
\date{}
\begin{document}
\maketitle
\begin{abstract}
Memory Networks have emerged as effective models to incorporate Knowledge Bases (KB) into neural networks. By storing KB embeddings into a memory component, these models can learn meaningful representations that are grounded to external knowledge. However, as the memory unit becomes full, the oldest memories are replaced by newer representations. 

In this paper, we question this approach and provide experimental evidence that conventional Memory Networks store highly correlated vectors during training. While increasing the memory size mitigates this problem, this also leads to overfitting as the memory stores a large number of training latent representations. To address these issues, we propose a novel regularization mechanism named \textit{memory dropout} which 1) Samples a single latent vector from a distribution of redundant memories. 2) Ages redundant memories thus increasing their probability of overwriting them during training. This fully differentiable technique allows us to achieve state-of-the-art response generation in the Stanford Multi-Turn Dialogue and Cambridge Restaurant datasets. 
\end{abstract}
\begin{figure*}[ht]
\vskip 0.01in
\centerline{\includegraphics[scale=0.5]{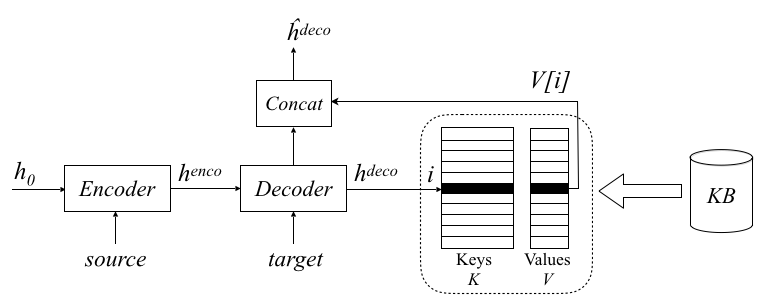}}
\caption{A  sequence-to-sequence model  augments its capacity by storing distinct latent representations of $h^{deco}$. }
\label{fig:mann}
\vskip 0.01in
\end{figure*}
\section{Introduction}
Having fluent conversations with a computer is one of the key challenges in machine learning~\cite{turing1950}. While users ask questions and provide commands, a  dialogue system responds based on its current model state. By augmenting this state with external KB information, the responses get grounded to a particular context. For example, answering questions like \textit{\textbf{What time is my dentist appointment?}} requires a model to access a calendar of events to predict which of the calendar tokens (e.g., names, dates, party) will be part of a valid response. Often, however, dialogue models generate responses that only leverage past dialogue history. This is because KBs are frequently stored in databases, so their access is not a differentiable operation.  The lack of seamless integration between neural networks and KBs limits the kind of natural conversations that people desire.  



An effective model to incorporate KB information into dialogue models is the Memory Network~\cite{key_value_networks}. This model encodes KB embeddings as key-value pairs and stores them into a memory component. The result is a conversational model that generates more fluent and informed responses~\cite{FDSM,Key_value_task_oriented_dialogue}. However, we have experimentally observed, over different datasets, that training Memory Networks leads to store highly correlated features. Researchers often increase the memory size to allow that more latent representations contribute to the prediction. But as the model capacity increases, it also does its overfitting. Despite being an important problem, there is no much work in regularizing the latent representations stored in external memories. In contrast to the \textit{dropout} technique, used to regularize deep neural networks ~\cite{dropout}, we propose the  \textit{memory dropout} mechanism to reduce overfitting in Memory Networks. 

Our technique works differently than the conventional \textit{dropout} technique as it takes advantage of the underlying properties of a memory component:  addressable embeddings and long-term storage. Rather than \textbf{\textit{immediately}} removing information with some probability as in~\cite{dropout}, we assign the oldest recorded age to redundant memories to increase the probability of overwriting them during training. This makes  \textit{memory dropout} a \textbf{\textit{delayed}} regularization mechanism.

The main contributions of our work are the following:
\begin{itemize}
\item We introduce a new regularization method that controls overfitting in Memory Networks. To our best knowledge, ours is the first work on regularizing these architectures.
\item We build a neural dialogue system that uses \textit{memory dropout} to incorporate a KB for automatic response generation. The results show that our technique can generate more fluent and accurate responses: an improvement of +$2.5$ and +$2.6$ BLEU points, and +$3.6\%$ and +$4.8\%$ Entity F1 scores, in the Stanford Multi-Turn Dialogue and Cambridge Restaurant datasets, respectively.
\end{itemize}
\section{Memory Augmented Neural Networks}
Memory Augmented Neural Networks (MANN) transforms inputs into latent vectors and store them into an external memory component, thus incrementing the overall model capacity. This capability is supported by a memory $M$ formed by arrays of keys $K$, values $V$, and age-related information $A$,  as introduced  in~\cite{rare_events}. 
\begin{equation}
M = (K, V, A)
\label{eq_memory}
\end{equation}
The memory module $M$ stores individual memories as key-value pairs. \textit{Keys} store learnable embeddings that work as local addresses for content-based lookups,  and \textit{values} store read-only vectors that provide context to those keys. 

Improving the quality of responses generated by a dialogue system requires to augment a sequence-to-sequence architecture with a memory module, as illustrated in Figure~\ref{fig:mann}. 
The new model stores longer the hidden representations learned during training. In more detail, an \textit{encoder} transforms the source sequence into the latent vector $h^{enco}$. Then, a \textit{decoder} takes both  this vector and the target sequence to generate the latent vector $h^{deco}$.  Before transforming $h^{deco}$ to an output, the decoder finds the best location in memory for incorporating this information. This is similar to executing a \textit{read} followed by a \textit{write} operation. 

We \textit{read} the memory by computing the Cosine similarity between  $h^{deco}$  and each key. Let $i$ be the index of the most similar key,
\begin{equation}
\begin{aligned}
i &= argmax_i(h^{deco} \cdot K[i])\\
\end{aligned}
\label{eq_i}
\end{equation}
The index $i$  indicates where to write in the memory module. In fact, the decoder \textit{writes} into the memory by updating the vector stored in $K[i]$ towards the direction of the current embedding $h^{deco}$.  
\begin{equation}
\begin{aligned}
K[i]  &= \frac{K[i] + h^{deco}}{||K[i + h^{deco}]||} 
\end{aligned}
\end{equation}
The index $i$  also indicates how to access a memory value $V[i]$. By concatenating $h^{deco}$ and $V[i]$ as a single vector, the decoder generates a memory-augmented latent vector $\hat{h}^{deco}$. In our case, $V[i]$ contains KB embeddings, as will discuss in Section~\ref{sec_memory_dropout}. 
\begin{equation}
\begin{aligned}
\hat{h}^{deco}  &= W[h^{deco} ; V[i]]
\end{aligned}
\end{equation}
When the memory component is full, Memory Networks follows a greedy strategy to replace old memories:  the decoder selects a random key and overwrites its content with probability $\epsilon$. We want to find a better policy for writing information in $M$.  
\section{Memory Dropout}
This section describes our main contribution, the \textit{memory dropout} neural model, which factors in the concept of \textit{age} to regularize external memories. To support our technique, we extend the memory definition in Equation~\ref{eq_memory} with array $S$ to also track the \textit{variance} of each key. Then, the final form of the memory module is as follows,
\begin{equation}
M = (K, V, A, S).
\label{eq_tuple}
\end{equation}
The memory component of a MANN gets saturated with highly correlated keys after consecutive training steps (see experiments in Section~\ref{sec_exp_memcorr}). Unfortunately, this issue produces overfitting (see experiments in Section~\ref{sec_exp_overfitting}).  Our solution is to learn a mathematical space in which redundant memories become older and are progressively forgotten. Core to this idea is the definition of the \textit{memory dropout} mechanism which takes place every time we write a latent vector $h$ to the memory. The full algorithm is presented in Algorithm~\ref{algo}.

To control the  effects of regularization, the $P$ most similar keys to $h$ define an updating neighborhood, $M' = (K', S')$. We find evidence that sampling a vector $h'$ from a distribution of redundant keys preserve the knowledge acquired during training. So, we sample $h'$ from a \textit{Gaussian Mixture Model} (GMM) parameterized by the locations and variances of the neighboring keys in $K'$.

More formally, the  updating region around $h$ is a subpopulation of latent vectors $h'$ that can be represented by a linear superposition of $P$ Gaussian components 
\begin{equation}
P(h') = \sum_{p=1}^{P} \pi_p \mathcal{N} (h'| \mu_p, \Sigma_p).
\label{eq_gaussian}
\end{equation}
Each Gaussian component is centered at a key ($\mu_p=k'_p$) with a covariance matrix given by $\Sigma_p=diag(s'_p)$, where $k'_p \in K'$ and $s'_p \in S'$. The variance $S'$ stores the uncertainty of each key in approximating $h$ and controls that large keys to not dominate the likelihood probability. The vector of probabilities $\pi=\{\pi_1, ..., \pi_P\}$ contains the mixing coefficients of the Gaussian components and it is defined as $\pi = Softmax(h \cdot K')$.
%
%

Sampling from the Gaussian Mixture Model requires to choose first the $j$-th Gaussian component under the categorical distribution $\pi$, and then to sample $h'$ from that mixture component.  
\begin{equation}
\begin{aligned}
j & \sim \pi \\
\quad\text{}
h' & \sim \mathcal{N} (h' | \mu_j, \Sigma_j).
\end{aligned}
\end{equation}
The vector $h'$ now characterizes the information of its neighbors. And we can write the current embedding $h$ by first computing the index $i$ of the most similar key in $K'$, as in Equation~\ref{eq_i}, and then updating the key $K[i]$ and variance $S[i]$ as follows,
\begin{equation}
\begin{aligned}
K[i]  &= \frac{h' + h}{||h' + h||} \\
\quad\text{}
S[i] &= (h - h')^2 \\
\end{aligned}
\end{equation}
While $A[i]$ is reset to zero to indicate recentness, the age of the other memories increases by one. This is to make older memories that are not frequently accessed during training. 

\begin{algorithm}
\caption{Write Memory Dropout (h, K, S, A)}
\begin{algorithmic}
\IF{$random() < \epsilon$} 
\STATE $//\ index\ of\ the\ oldest\ memory$
\STATE $i \leftarrow argmax_i(A[i])$
\STATE $//\ overwrite\ old\ key$
\STATE $K[i]  \leftarrow h$
\STATE $//\ reset\ variance\ and\ age$
\STATE $S[i], A[i] \leftarrow (0, 0)$
\ELSE{}
\STATE $//\ to\ lookup\ based\ on\ content$
\STATE $attention \leftarrow Softmax(h \cdot K)$
\STATE $//\ index\ of\ P\ neighbors\ around\ h$
\STATE $i\_knn  \leftarrow Sort(attention)[0\mathbin{:}P]$
\STATE $neigh\_h \leftarrow K[i\_knn]$
\STATE $//\ sampling\ a\ new\ latent\ vector\ h'$
\STATE $h' \sim GMM(neigh\_h)$
\STATE $//\ index\ of\ the\ most\ similar\ key$
\STATE $i \leftarrow i\_knn[0]$
\STATE $//\ key\ update$
\STATE $K[i] \leftarrow (h' + h)/||h' + h||$
\STATE $//\ penalize\ redundant\ memories$
\STATE $A[i\_knn] \leftarrow max(A)$
\STATE $//\ update\ variance\ and\ age$
\STATE $S[i], A[i] \leftarrow (h - h')^2, 0$
\ENDIF
\STATE $//\ all\ memories\ get\ older$
\STATE $A \leftarrow A + 1$
\end{algorithmic}
\label{algo}
\end{algorithm}

This approach has several advantages over the greedy strategy to maintain memories used in Memory Networks. First, each writing to the memory does not remove old memories immediately but it gives them the chance to contribute to the prediction later. Second, directly removing the oldest memories is inefficient, due to the strong correlations between keys; sampling from their distribution preserves information and maximizing their age increases the chance of replacing an entire region with recent information later. This improves the diversity of information stored in the memory component (see experiments in Section~\ref{sec_exp_memcorr}).
%

%
%
\section{Using Memory Dropout}
\label{sec_memory_dropout}
\begin{figure*}[ht]
\vskip -0.01in
\centering
\centerline{\includegraphics[scale=0.45]{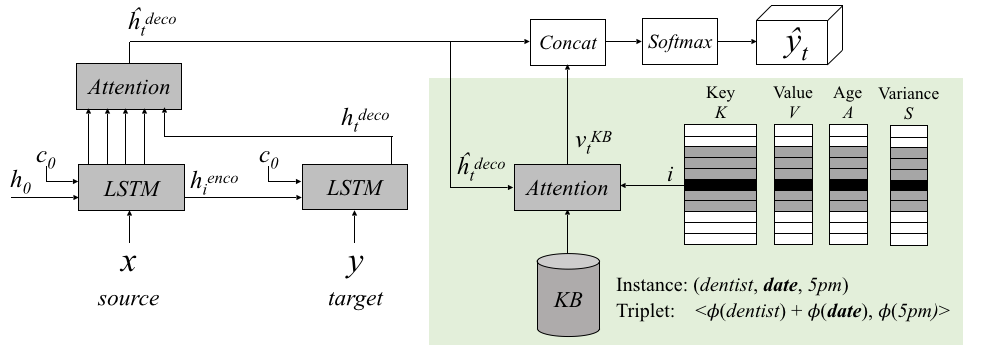}}
\caption{Architecture of a neural dialogue model that incorporates a KB. The $i$-th decoding step of the word $\hat{y_i}$ attends a memory module that uses \textit{memory dropout} for memory regularization.}
\label{fig:seq2seqKB}
\vskip -0.01in
\end{figure*}

We now study \textit{memory dropout} in a realistic scenario: as the memory unit of a dialogue system that generates responses grounded to a Knowledge Base. Dialogue consists of multiple turns grounded to a common KB. Our goal is to learn KB embeddings (e.g., a calendar of events) that guide the accurate generation of responses. This task is challenging because neural dialogue agents struggle to interface with the structured representations of KBs. 
Our proposed architecture consists of a Sequence-to-Sequence model to represent the dialogue history, and a memory component to represent the KB, as shown in Figure~\ref{fig:seq2seqKB}.  All updates on the external memory follow the \textit{memory dropout} mechanism described in Algorithm~\ref{algo}.

To encode KB information, we decompose its tabular format into a set of triplets that contain the following information  \textit{(subject, \textbf{relation}, object)}, as originally introduced by ~\cite{Key_value_task_oriented_dialogue}. For example, the following entry  in the KB represents a dentist appointment

\begin{center}
\begin{tabular}{ |c|c|c|c| } 
 \hline
\textbf{event} & \textbf{date} & \textbf{time} & \textbf{party}\\
 \hline
dentist & the 19th & 5pm & Mike\\
 \hline
\end{tabular}
\end{center}

which can expand to 12 different triplets by considering each column name (e.g., event, date, time, party) as a \textbf{relation}:
\medskip 
 \begin{adjustwidth}{-0.26cm}{}
\begin{tabular}{ |l|l|} 
 \hline
\textit{(\textit{dentist}, \textit{\textbf{time}}, \textit{5pm})} & \textit{(\textit{the 19th}, \textit{\textbf{event}}, \textit{dentist})}\\
 \textit{(\textit{dentist}, \textit{\textbf{party}}, \textit{Mike})} & \textit{(\textit{the 19th}, \textit{\textbf{party}}, \textit{Mike})}\\
\textit{(\textit{dentist}, \textit{\textbf{date}}, \textit{the 19th})}  & \textit{(\textit{the 19th}, \textit{\textbf{time}}, \textit{5pm})}\\
\textit{(\textit{Mike}, \textit{\textbf{time}}, \textit{5pm})} &  \textit{(\textit{5pm}, \textit{\textbf{event}}, \textit{dentist})}\\
\textit{(\textit{Mike}, \textit{\textbf{event}}, \textit{dentist})} &  \textit{(\textit{5pm}, \textit{\textbf{date}}, \textit{the 19th})}\\
\textit{(\textit{Mike}, \textit{\textbf{date}}, \textit{the 19th})} &  \textit{(\textit{5pm}, \textit{\textbf{party}}, \textit{Mike})}\\
 \hline
\end{tabular}
\end{adjustwidth}
\medskip 
By considering subject and \textbf{relation} elements in the \textit{key} and the object in the \textit{value}, these triplets generate key-value pairs.
\begin{equation}
\langle \norm{ \phi^{emb}(\textit{subject}) + \phi^{emb}( \textbf{\textit{relation}} ) }, \phi^{emb}(\textit{object}) \rangle
\label{eq_keyvalue}
\end{equation}
here $\phi^{emb}$ is the same word embedding that maps input tokens to fixed-dimensional vectors in the model\footnote{We use the GloVe word embedding of 300 dimensions}. 

During training, key-value embeddings feed the memory module and  sequences of token embeddings from a dialogue feed the sequence-to-sequence model. For example, token embeddings of the input sequence 
$$x = \text{\textbf{\textit{``What \underline{time} is my \underline{dentist} appointment?''}}}$$  likely activate the following memory  
\begin{equation*}
\langle \norm{ \phi^{emb}(\underline{dentist}) + \phi^{emb}( \textbf{\underline{\textit{time}}} ) },\ \phi^{emb}(\textit{5pm}) \rangle
\end{equation*}
and its corresponding \textit{value}, $\phi^{emb}(\textit{5pm})$, is returned to define an output latent vector.

To model dialogue information, we use an encoder-decoder architecture that employs two LSTMs~\cite{seq2seq}.  The encoder LSTM receives an source sequence $x$ and outputs a hidden vector $h_i^{enco}$ for each input token $x_i$ following the recurrence,
\begin{equation}
\begin{aligned}
h^{enco}_{i} &= LSTM( \phi^{emb}(x_i),\ h^{enco}_{i-1})\\ 
\end{aligned}
\end{equation} 
At every timestep $t$, the LSTM decoder gets the current target token $y_{t}$ and the previous hidden state $h^{deco}_{t-1}$ to generate the hidden state $h^{deco}_{t}$.  Then, the decoder uses $h^{deco}_{t}$ to  attend to every hidden vector  of the encoder $h_i^{enco}$ and weights their contribution in generating the  final decoder hidden state $\hat{h}^{deco}_t$, 

\begin{equation}
\begin{aligned}
h^{deco}_{t} &= LSTM( \phi^{emb}(y_t), h^{deco}_{t-1})\\ 
\hat{h}^{deco}_t &= Softmax(h^{deco}_t \cdot h^{enco})\ h^{deco}_t.
\end{aligned}
\end{equation} 
Attending the memory keys with $\hat{h}^{deco}_t$  induces an unnormalized distribution from which we pick the index $i$ of the most similar key. Instead of using $\hat{h}^{deco}_t$ to predict an output, we augment its information with the KB value  $v^{KB}_t$. 
\begin{equation}
\begin{aligned}
i &= argmax_i(\hat{h}^{deco}_t \cdot K[i])\ \\
v^{KB}_t &= V[i] \\
\end{aligned}
\end{equation}
The output vector $o_t$ concatenates both dialogue $\hat{h}^{deco}_t $ and KB representation $v^{KB}_t$  as a single vector which is mapped into the vocabulary space for a Softmax function to decode the current token $\hat{y}_t$,
\begin{equation}
\begin{aligned}
o_t &= W[\hat{h}^{deco}_t;\ v^{KB}_t] \\
\hat{y}_t &= Softmax(o_t) 
\end{aligned}
\end{equation}
Naturally, the objective function is to minimize the cross-entropy between the actual and generated responses:
\begin{equation}
J(\theta) = -\sum_{j=1}^{N} \sum_{k=1}^{L_j} \sum_{t=1}^{m} y_{t}^{j, k}\ log\ p(\hat{y}_{t}^{j, k})
\label{eq_loss}
\end{equation}
\noindent where $N$ is the number of dialogues, $L_j$ is the number of turns in the $j^{th}$ dialogue, $m$ is the length of the generated response. And $y_{t}^{j, k}$ is $t$-th word in the actual response in one-hot encoding representation.

\section{Experimental Design}
\label{label_experiments}

We evaluate the effectiveness of \textit{memory dropout} in four aspects: fluency and task completion (Section~\ref{sec_exp_response}), management of redundant keys (Section \ref{sec_exp_memcorr}), regularization power (Section \ref{sec_exp_overfitting}), and dependency on the memory size (Section \ref{sec_exp_memsize}). 

\subsection{Datasets}
\label{sec_exp_dataset}
We tested our proposed method in the \textit{Stanford Multi-Turn Dialogue}\footnote{\url{https://nlp.stanford.edu/blog/a-new-multi-turn-multi-domain-task-oriented-dialogue-dataset/}}  (SMTD) dataset~\cite{Key_value_task_oriented_dialogue} and the \textit{Cambridge Restaurant}\footnote{\url{https://www.repository.cam.ac.uk/handle/1810/260970}} (CamRest) dataset~\cite{CamRest}. Both datasets contain multi-turn dialogues collected using a Wizard-of-Oz scheme on Amazon Mechanical Turk. Workers have two potential roles: user or system and carry out fluent conversations to complete a task.  

The SMTD dataset consists of 3,031 dialogues in the domain of an in-car assistant, which provides automatic responses grounded to a personalized KB, known only to the in-car assistant. The entries of the KB encode information for satisfying a query formulated by the driver. There are three types of KBs:  \textit{schedule of events}, weekly \textit{weather forecast}, and information for point-of-interest \textit{navigation}. The CamRest dataset consists of 680 dialogues in the domain of restaurant reservations. Following previous works, the SMTD and CamRest datasets are split 8:1:1 and 3:1:1 into training, developing, and testing sets, respectively. The vocabulary size is 14,000 for SMTD and 800 for CamRest.

\begin{table*}[t]
\begin{center}
\begin{small}
\begin{sc}
\begin{tabular}{lccccccc}
\cmidrule(lr){2-6}\cmidrule(lr){7-8}
& \multicolumn{5}{c}{SMTD} & \multicolumn{2}{c}{CamRest} \\
\cmidrule(lr){2-6}\cmidrule(lr){7-8}

Model & Sche.F1 & Weat.F1& Navi.F1  & BLEU & Ent.F1 & BLEU & Ent.F1\\
\midrule
Seq2Seq+Attn. 		& 35.4$\pm$0.3		&	44.6$\pm$0.8	& 23.2$\pm$1.2		& 12.3$\pm$0.2		& 32.4$\pm$0.7 	&	11.5$\pm$0.1	& 41.5$\pm$0.5	\\
KVRN 				& 63.1$\pm$0.3		& 	47.4$\pm$0.6	& 42.1$\pm$1.1		& 13.8$\pm$0.2 		& 49.2$\pm$0.6	& 	16.8$\pm$0.1	& 59.8$\pm$0.6	\\
FDSM/St 			& 77.9$\pm$0.3	 	&	74.5$\pm$0.6	& 70.2$\pm$1.0		& 20.1$\pm$0.3 		& 80.1$\pm$0.8 	& 	24.2$\pm$0.2	& 83.6$\pm$0.8 \\
\midrule
MANN/KB 			& 38.7$\pm$0.4 		& 	46.6$\pm$0.5	& 26.6$\pm$0.9 		& 12.8$
\pm$0.2  	& 35.6$\pm$0.7  &	12.1$\pm$0.2	& 44.3$\pm$0.8	\\
MANN+MD/Enco 		& 68.8$\pm$0.5		&	60.3$\pm$0.8	& 56.4$\pm$0.9		& 14.0$\pm$0.3 		& 58.1$\pm$0.9 	&	18.4$\pm$0.3	& 69.1$\pm$0.8	\\
MANN/MD				& 75.5$\pm$0.3		& 	70.7$\pm$0.6 	& 68.1$\pm$1.1 		& 18.8$\pm$0.3  	& 78.2$\pm$0.8  &	23.7$\pm$0.3	& 80.8$\pm$0.9	\\
\midrule
\textbf{MANN+MD} 	& \textbf{81.8$\pm$0.3} &  \textbf{77.2$\pm$0.4}	& \textbf{74.6$\pm$0.6} & \textbf{21.3$\pm$0.3}  & \textbf{81.8$\pm$0.6} 	&  \textbf{26.3$\pm$0.2}	& \textbf{85.6$\pm$0.8}	\\
\bottomrule
\end{tabular}
\end{sc}
\end{small}
\end{center}
\caption{BLEU and Entity F1 scores for the SMTD and CamRest datasets. Results are averaged after 10 independent training processes.}
\label{table:exp_bleu}
\end{table*}

\subsection{Training Settings}
\label{label_settings}
For all the experiments, we use the GloVe word embedding of size 300~\cite{pennington-etal-2014-glove}. The encoder and decoder are bidirectional LSTMs of 3-layers with a state size of 256. The dimension of all hidden layers is 256. We initialized all weights with samples from a normal distribution with mean zero and a standard deviation of 0.0001. We minimize the loss function in Equation~\ref{eq_loss} with the Adam optimizer~\cite{ADAM}, with a learning rate of $0.005$. We also applied regular dropout after the input and output of each LSTM with keep probability of $95.0\%$ for SMTD and $50.0\%$ for CamRest. We identified the overwrite probability $\epsilon$ by random search in $[0.01, 5]$ and fixed it at $0.1$ after evaluating in the held-out validation dataset.

\subsection{Baselines}
\label{sec_exp_baselines}
We compare our model \textbf{Memory Augmented Neural Network with Memory Dropout} (MANN+MD) to three baseline models:

1) \textbf{Seq2Seq+Attention}~\cite{seq2seq_attention}:  An encoder-decoder architecture that maps between sequences with minimal assumptions on the input structure. This method attends to parts of the dialogue history to predict a target word. 2) \textbf{Key-Value Retrieval Network} (KVRN)~\cite{Key_value_task_oriented_dialogue}: A memory augmented encoder-decoder that computes attention over the memory keys of a KB. 3) \textbf{FDSM/St}~\cite{FDSM}: A recent sequence-to-sequence approach with a copy-augmented sequential decoder that deals with unknown values in the conversation and does not consider explicit state tracking in the response decoding. Only baselines 2 and 3 have access to a KB.

We also perform the following ablation studies to validate the necessity of \textit{memory dropout}:

1) \textbf{MANN/MD}: The same sequence-to-sequence architecture shown in Figure~\ref{fig:seq2seqKB} using a memory component but not memory dropout. 2)  \textbf{MANN/KB}:  We remove the attention to the KB and generate responses by only attending to the dialogue.  3) \textbf{MANN/Enco}: We remove the attention to the encoder so the decoder only attends to the KB. This prevents the dialogue history from directly conditioning the generation of responses.

\subsection{Results on System Response Generation}
\label{sec_exp_response}

Evaluating dialogue systems is challenging because a trained model can generate free-form responses. As in \cite{Key_value_task_oriented_dialogue}, we employ two metrics to quantify the performance of a dialogue agent grounded to a knowledge base. 

\begin{itemize}
\item\textbf{BLEU}~\cite{bleu}: A metric of fluency that looks at $n$-gram precisions for $n=1,2,3,4$ comparing exact matches of words between model and human responses. 
\item \textbf{Entity F1} : A metric of task completion that measures the correct retrieval of entities expected to be present in a generated response. 
\end{itemize}

As shown in Table~\ref{table:exp_bleu}, we found that MANN+MD improves both dialogue fluency (BLEU) and task completion (Entity F1 score) in comparison to all other models that also use a KB such as KVRN and FDSM/St. Compared to the recent FDSM/St, both models interact with a KB during decoding but MANN+MD stores latent representations. This end-to-end approach learns powerful representations that approximate an implicit dialogue state that is needed to predict responses during a dialogue. The Seq2Seq+Attn model shows the worst performance as it only relies on dialogue history to answer questions that often require some knowledge of the world.

Ablation studies validate the necessity of incorporating a KB in the generation of responses. For example, MANN/KB does not attend to the KB and performs worse than MANN/MD and MANN+MD in both the SMTD and CamRest datasets. Other components of the architecture are also important. For example, not attending to encoder outputs (MANN+MD/Enco)  has a less severe effect than not attending the KB. Also, comparing two identical architectures MANN+MD and MANN/MD, i.e. using MD or not,  results in a difference in the F1 Entity score of $3.6\%$ for the SMTD and $4.8\%$ for the CamRest dataset. 


Using a memory component that penalizes redundant keys could explain the gains obtained by MANN+MD. To test this hypothesis, we study now the correlation of keys in Section~\ref{sec_exp_memcorr}.

\subsection{Results on Correlated Memories}
\label{sec_exp_memcorr}

We found that keys become redundant as training progresses. To observe this effect, we compute the aggregated Pearson correlation coefficient between pairs of memory keys at different training steps. 

\begin{figure*}
\begin{subfigure}{.53\textwidth}
\centering
\includegraphics[width=1.05\linewidth]{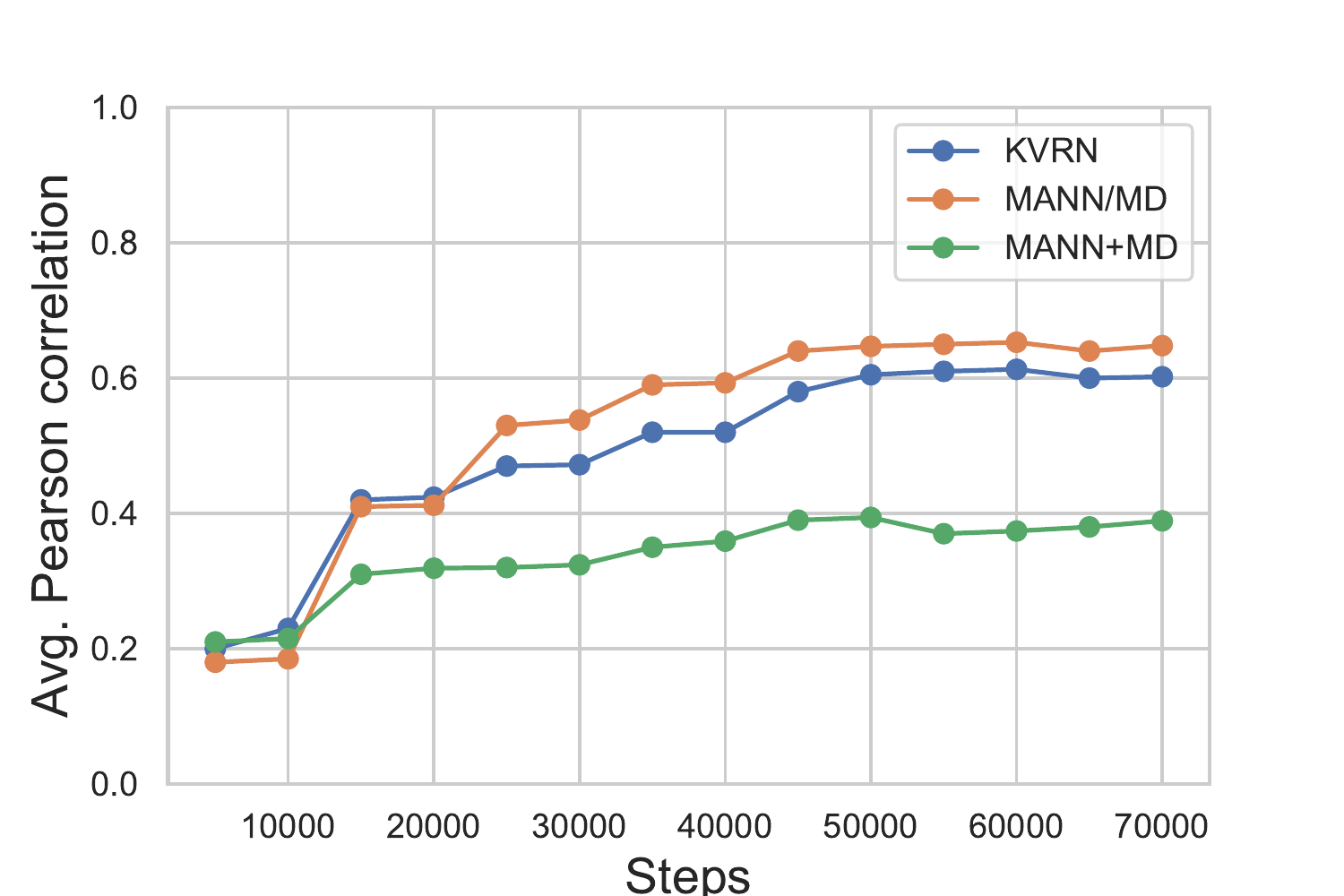}
\caption{SMTD}
\label{fig:sfig1}
\end{subfigure}%
\begin{subfigure}{.53\textwidth}
\centering
\includegraphics[width=1.05\linewidth]{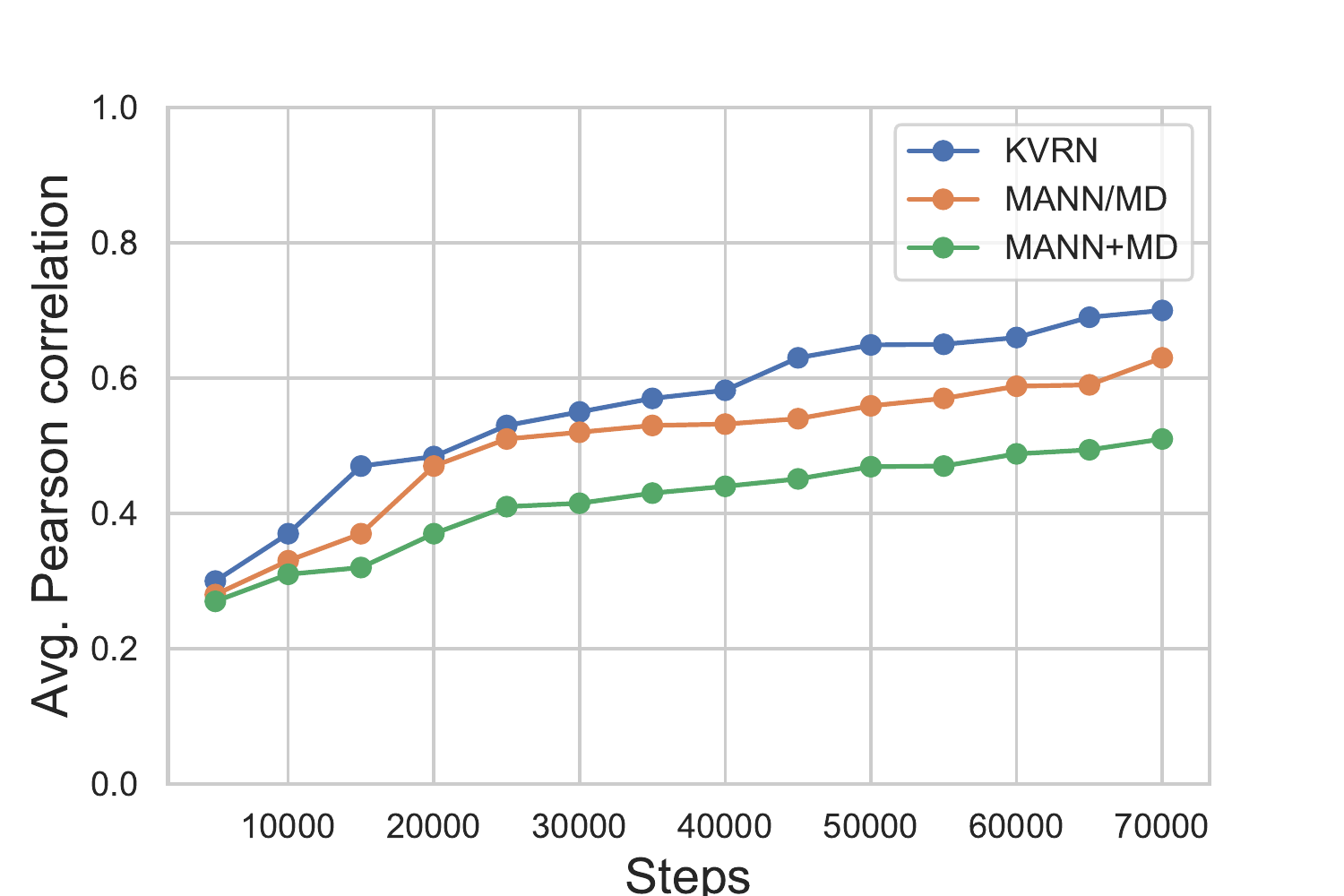}
\caption{CamRest}
\label{fig:sfig2}
\end{subfigure}
\caption{Average Pearson correlation values between pairs of keys at different training steps. As training progresses, redundant keys are allocated to the external memory. Lower values are preferred.}
\label{fig:correlations}
\end{figure*}

Figure~\ref{fig:correlations} compare the degree of linear correlations in the memory networks studied in this paper. We can see that all models initially show small correlation values as keys are randomly initialized. Then, both MANN and KVRN show larger correlation values indicating a strong linear dependency between keys. In contrast, MANN+MD shows correlation values that do not increase at the same rate as the other methods for both the SMTD and CamRest datasets. We empirically conclude that the use of \textit{memory dropout} penalizes the presence of redundant keys which results in storing more diverse latent representations in memory.

\subsection{Results on Overfitting Reduction}
\label{sec_exp_overfitting}
 In an attempt to isolate the contribution of \textit{memory dropout} in dealing with overfitting, we disable \textit{dropout} in the input and output of the encoder-decoder architecture. Then, we evaluate the Entity F1 scores of the MANN/MD and MANN+MD models considering different neighborhood sizes.

\begin{figure*}
\begin{subfigure}{.53\textwidth}
\centering
\includegraphics[width=1.0\linewidth]{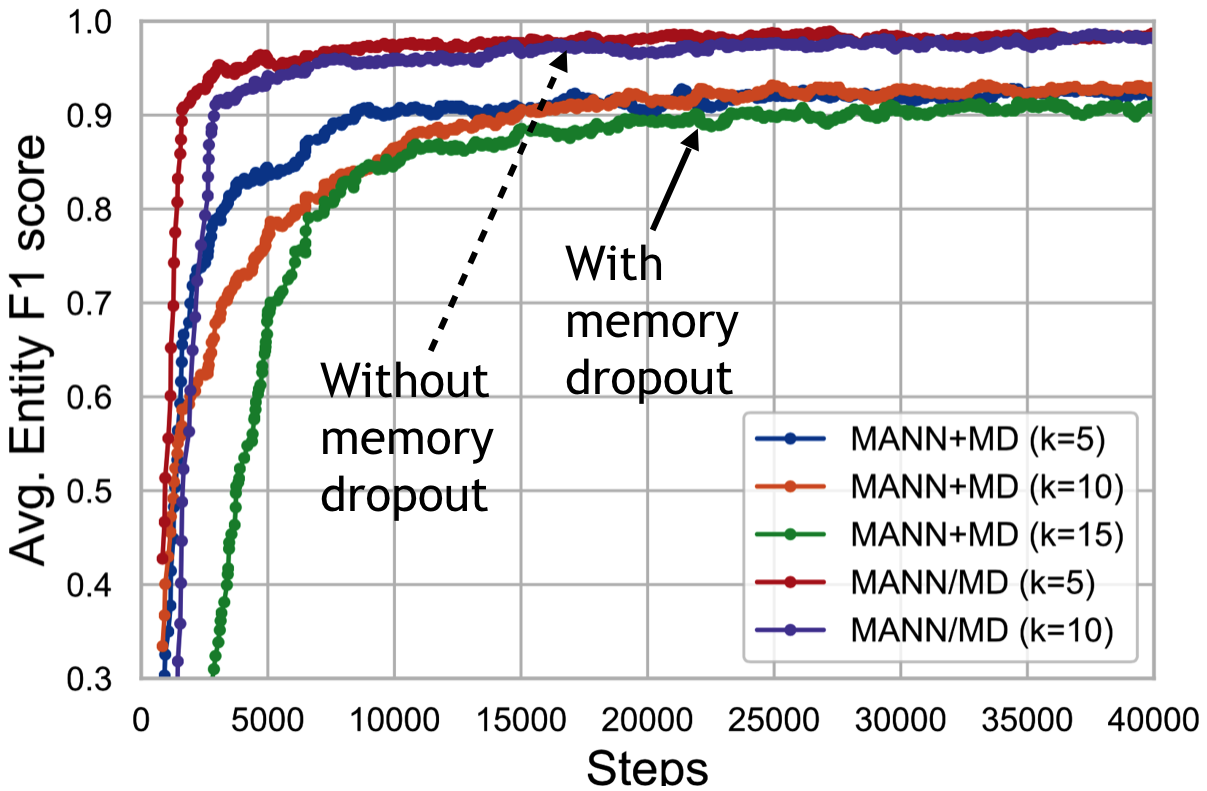}
\caption{Training Entity F1}
\label{fig:sfig1}
\end{subfigure}%
\begin{subfigure}{.53\textwidth}
\centering
\includegraphics[width=1.0\linewidth]{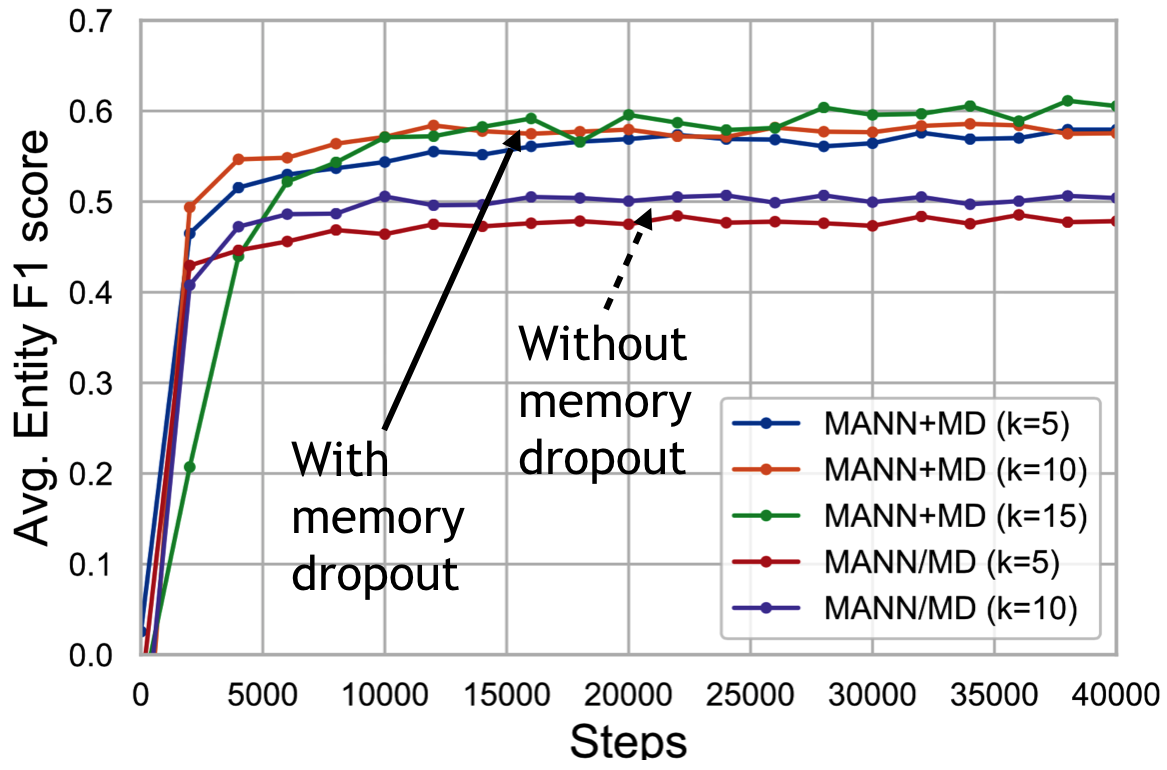}
\caption{Testing Entity F1}
\label{}
\end{subfigure}
\caption{Entity F1 scores considering the use of \textit{memory dropout} in the MANN model for the SMTD dataset.}
\label{fig:overfitting}
\end{figure*}

Figure~\ref{fig:overfitting} shows two groups of behavior in each plot. During training, not using \textit{memory dropout} (MANN/MD) leads to obtain higher Entity F1 scores, a reasonable outcome considering that not regularizer is present in the model. During testing, MANN/MD shows lower Entity F1 scores indicating that the model overfits to its training dataset and has problems generalizing in the testing dataset. On the other hand, the use of \textit{memory dropout} (MANN+MD) provides a more conservative performance during training but better Entity F1 scores during testing resulting in an average improvement of $10\%$. Testing with different neighborhood sizes (from 5 to 15 elements) shows two groups of behavior according to whether a model uses \textit{memory dropout} or not.

\begin{figure*}
\begin{subfigure}{.53\textwidth}
\centering
\includegraphics[width=1.05\linewidth]{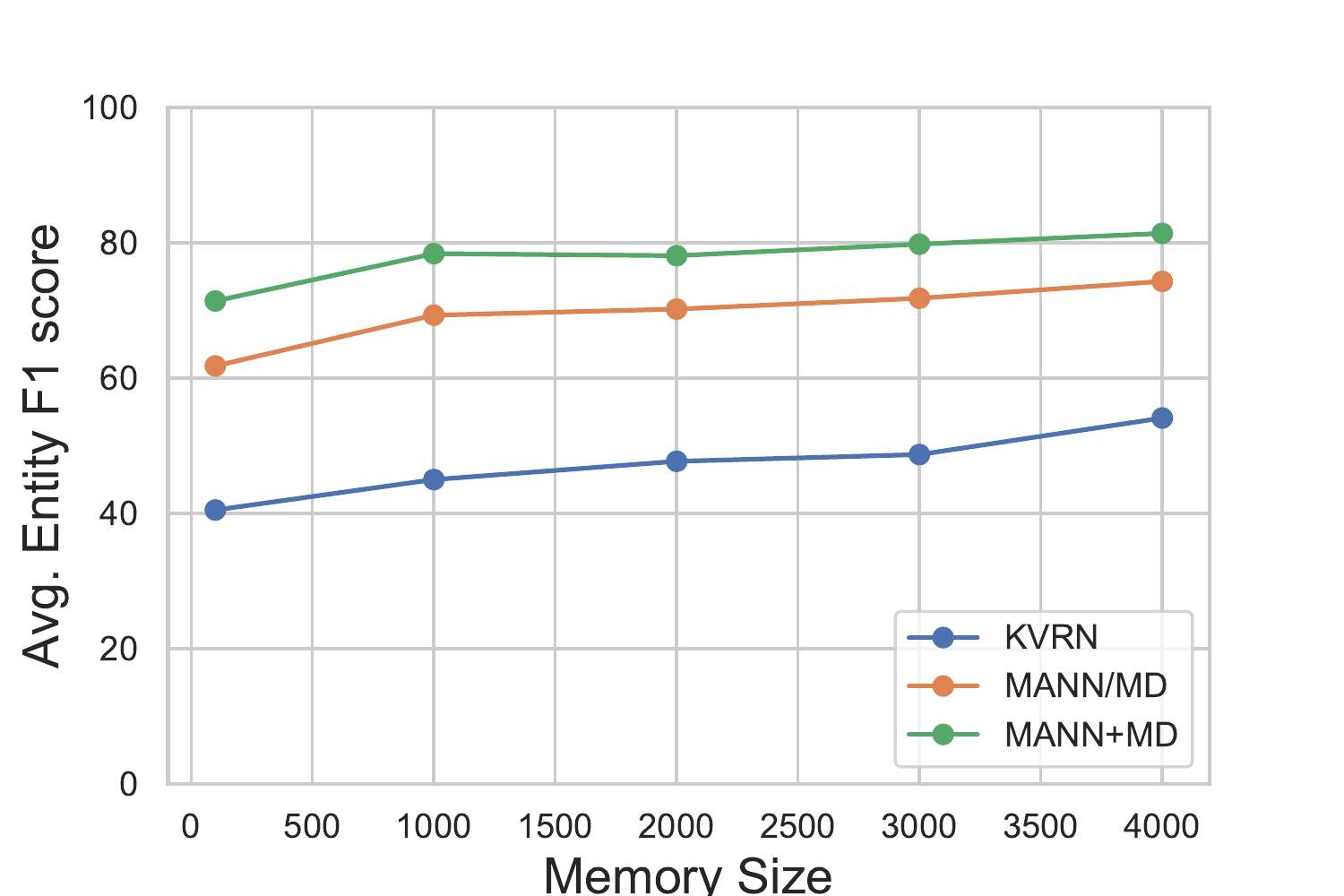}
\caption{SMTD}
\label{fig:sfig1}
\end{subfigure}%
\begin{subfigure}{.53\textwidth}
\centering
\includegraphics[width=1.05\linewidth]{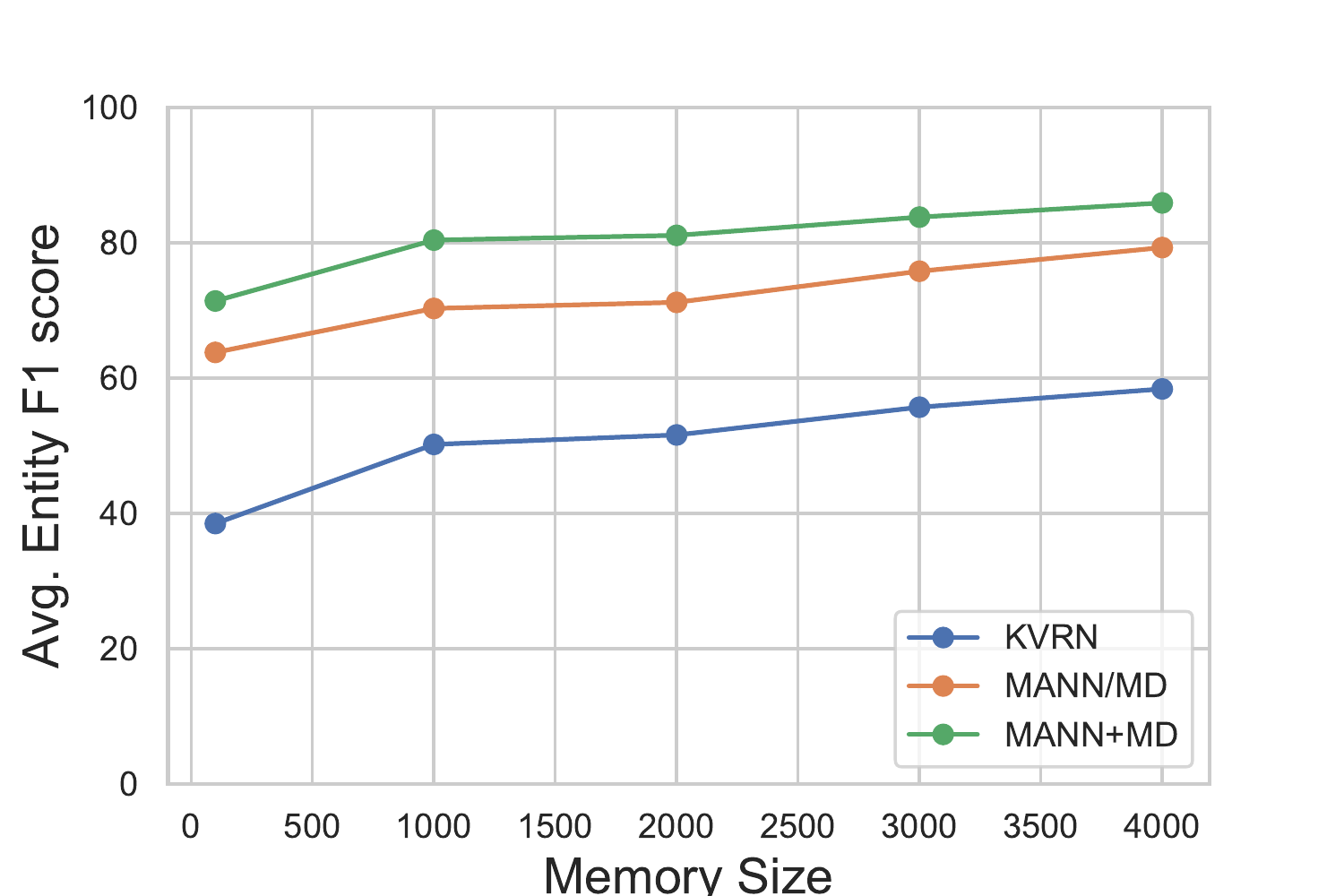}
\vskip -0.01in
\caption{CamRest}
\label{fig:sfig2}
\end{subfigure}
\caption{Entity F1 scores on the testing dataset given different memory sizes. Higher values are preferred.}
\label{fig:memory_size}
\end{figure*}

\subsection{Results on the Effect of Memory Size}
\label{sec_exp_memsize}
One side-effect of using Memory Networks is the need for larger memories (number of keys) to accommodate the redundant activations generated during training. To evaluate the usefulness of using large memories for encoding a KB, we compare the Entity F1 score for all the models that use an external memory considering different memory sizes. 

Figure~\ref{fig:memory_size} shows that while MANN/MD and KVRN show higher scores when increasing the memory size, MANN/MD outperforms KVRN.  Adding \textit{memory dropout} (MANN+MD) increases the Entity F1 scores and results in the most accurate method across all memory configurations. As seen in experiments in Section~\ref{sec_exp_memcorr} (memory correlations) and Section~\ref{sec_exp_memsize} (memory size), using \textit{memory dropout} requires smaller memories as less redundant memories are stored in memory which translates into higher levels of accuracy.

\section{Related Work}
\label{label_related}
Memory Networks consider a neural encoder to store the hidden states of the model into an external differentiable memory~\cite{Augmented}. 
Some representative examples include~\cite{rare_events}, which also uses external memory to extend the capacity of the architecture. Also, Neural Turing Machines (NTM)~\cite{graves2014neural} are differentiable architectures allowing efficient training via associative recall for learning different sequential patterns. In this paper, we extend the key-value architecture introduced in~\cite{rare_events} because of its simplicity and proved results in learning sequence-to-sequence models. 

Deep neural models have been used to train dialogue agents~\cite{Rojas-BarahonaG17}. More recent architectures use an external memory to encode a KB, ~\cite{Key_value_task_oriented_dialogue}. Although the key-value architecture of this system incorporates domain-specific knowledge, it overfits the training dataset impacting model accuracy. Our model contrasts with this work on designing a memory strategy that deals directly with overfitting while requiring smaller memory sizes. 

The regularization of neural networks is an important problem that affects learning deep models. Recently~\cite{regularize_transitions} proposes the regularization of state transitions in a recurrent neural network, but the notion of memory is still internal and individual memories cannot be addressed. The popular dropout technique works in hidden layers as a form of model averaging~\cite{dropout}. In contrast to~\cite{dropout}, our \textit{memory dropout} is a delayed (age-sensitive) regularization mechanism that works at the level of memories and not in individual activations. To the best of our knowledge, ours is the first work that addresses the regularization of Memory Networks and proves its effectivity in a challenging task such as automatic response generation.

\section{Conclusions}
\textit{Memory dropout} is a technique that breaks co-adapting memories built during backpropagation in memory augmented neural networks. While conventional dropout regularizes at the level of individual activations, our \textit{memory dropout} deals with latent representations stored in the form of keys in the discrete locations of external memory. This resembles some areas of the brain that are also content-addressable and sensitive to semantic information~\cite{Sparse}. 

Central to this technique is the idea that \textit{age} and \textit{uncertainty} play important roles when regularizing memories that are persistent across training steps. By doing this, we obtain state-of-the-art BLEU and Entity F1 scores and dialogue systems that generate more fluent responses.

\bibliographystyle{acl_natbib}
\bibliography{biblio}

\begin{thebibliography}{17}
\expandafter\ifx\csname natexlab\endcsname\relax\def\natexlab#1{#1}\fi

\bibitem[{Bahdanau et~al.(2014)Bahdanau, Cho, and Bengio}]{Augmented}
Dzmitry Bahdanau, Kyunghyun Cho, and Yoshua Bengio. 2014.
\newblock Neural machine translation by jointly learning to align and
  translate.
\newblock \emph{CoRR}, abs/1409.0473.

\bibitem[{Bahdanau et~al.(2015)Bahdanau, Cho, and Bengio}]{seq2seq_attention}
Dzmitry Bahdanau, Kyunghyun Cho, and Yoshua Bengio. 2015.
\newblock Neural machine translation by jointly learning to align and
  translate.
\newblock In \emph{3rd International Conference on Learning Representations,
  {ICLR} 2015, San Diego, CA, USA, May 7-9, 2015, Conference Track
  Proceedings}.

\bibitem[{Eric et~al.(2017)Eric, Krishnan, Charette, and
  Manning}]{Key_value_task_oriented_dialogue}
Mihail Eric, Lakshmi Krishnan, Francois Charette, and Christopher~D. Manning.
  2017.
\newblock Key-value retrieval networks for task-oriented dialogue.
\newblock In \emph{Proceedings of the 18th Annual SIGdial Meeting on Discourse
  and Dialogue}. Association for Computational Linguistics.

\bibitem[{Graves et~al.(2014)Graves, Wayne, and Danihelka}]{graves2014neural}
Alex Graves, Greg Wayne, and Ivo Danihelka. 2014.
\newblock \href {http://arxiv.org/abs/1410.5401} {Neural turing machines}.
\newblock Cite arxiv:1410.5401.

\bibitem[{Kaiser et~al.(2017)Kaiser, Nachum, Roy, and Bengio}]{rare_events}
Lukasz Kaiser, Ofir Nachum, Aurko Roy, and Samy Bengio. 2017.
\newblock Learning to remember rare events.

\bibitem[{Kingma and Ba(2014)}]{ADAM}
Diederik~P. Kingma and Jimmy Ba. 2014.
\newblock Adam: A method for stochastic optimization.
\newblock CoRR.

\bibitem[{Miller et~al.(2016)Miller, Fisch, Dodge, Karimi, Bordes, and
  Weston}]{key_value_networks}
Alexander Miller, Adam Fisch, Jesse Dodge, Amir-Hossein Karimi, Antoine Bordes,
  and Jason Weston. 2016.
\newblock \href {https://doi.org/10.18653/v1/D16-1147} {Key-value memory
  networks for directly reading documents}.
\newblock In \emph{Proceedings of the 2016 Conference on Empirical Methods in
  Natural Language Processing}, pages 1400--1409, Austin, Texas. Association
  for Computational Linguistics.

\bibitem[{Papineni et~al.(2002)Papineni, Roukos, Ward, and Zhu}]{bleu}
Kishore Papineni, Salim Roukos, Todd Ward, and Wei-Jing Zhu. 2002.
\newblock \href {https://doi.org/10.3115/1073083.1073135} {Bleu: a method for
  automatic evaluation of machine translation}.
\newblock In \emph{Proceedings of 40th Annual Meeting of the Association for
  Computational Linguistics}, pages 311--318, Philadelphia, Pennsylvania, USA.
  Association for Computational Linguistics.

\bibitem[{Pennington et~al.(2014)Pennington, Socher, and
  Manning}]{pennington-etal-2014-glove}
Jeffrey Pennington, Richard Socher, and Christopher Manning. 2014.
\newblock {G}love: Global vectors for word representation.
\newblock In \emph{Proceedings of the 2014 Conference on Empirical Methods in
  Natural Language Processing ({EMNLP})}.

\bibitem[{Rojas-Barahona et~al.(2017)Rojas-Barahona, Gasic, Mrksic, Su, Ultes,
  Wen, Young, and Vandyke}]{Rojas-BarahonaG17}
Lina~Maria Rojas-Barahona, Milica Gasic, Nikola Mrksic, Pei-Hao Su, Stefan
  Ultes, Tsung-Hsien Wen, Steve~J. Young, and David Vandyke. 2017.
\newblock A network-based end-to-end trainable task-oriented dialogue system.
\newblock Association for Computational Linguistics.

\bibitem[{Shu et~al.(2019)Shu, Molino, Namazifar, Xu, Liu, Zheng, and
  Tur}]{FDSM}
Lei Shu, Piero Molino, Mahdi Namazifar, Hu~Xu, Bing Liu, Huaixiu Zheng, and
  Gokhan Tur. 2019.
\newblock Flexibly-structured model for task-oriented dialogues.
\newblock In \emph{Proceedings of the 20th Annual SIGdial Meeting on Discourse
  and Dialogue}, Stockholm, Sweden. Association for Computational Linguistics.

\bibitem[{Srivastava et~al.(2014)Srivastava, Hinton, Krizhevsky, Sutskever, and
  Salakhutdinov}]{dropout}
Nitish Srivastava, Geoffrey Hinton, Alex Krizhevsky, Ilya Sutskever, and Ruslan
  Salakhutdinov. 2014.
\newblock Dropout: A simple way to prevent neural networks from overfitting.
\newblock \emph{J. Mach. Learn. Res.}, 15(1):1929--1958.

\bibitem[{Sutskever et~al.(2014)Sutskever, Vinyals, and Le}]{seq2seq}
Ilya Sutskever, Oriol Vinyals, and Quoc~V Le. 2014.
\newblock Sequence to sequence learning with neural networks.
\newblock In Z.~Ghahramani, M.~Welling, C.~Cortes, N.~D. Lawrence, and K.~Q.
  Weinberger, editors, \emph{Advances in Neural Information Processing Systems
  27}, pages 3104--3112. Curran Associates, Inc.

\bibitem[{Turing(1950)}]{turing1950}
A.~M. Turing. 1950.
\newblock Computing machinery and intelligence.
\newblock \emph{Mind}, 59:433--460.

\bibitem[{Wang and Niepert(2019)}]{regularize_transitions}
Cheng Wang and Mathias Niepert. 2019.
\newblock State-regularized recurrent neural networks.

\bibitem[{Wen et~al.(2017)Wen, Vandyke, Mrk{\v{s}}i{\'c}, Ga{\v{s}}i{\'c},
  Rojas-Barahona, Su, Ultes, and Young}]{CamRest}
Tsung-Hsien Wen, David Vandyke, Nikola Mrk{\v{s}}i{\'c}, Milica
  Ga{\v{s}}i{\'c}, Lina~M. Rojas-Barahona, Pei-Hao Su, Stefan Ultes, and Steve
  Young. 2017.
\newblock A network-based end-to-end trainable task-oriented dialogue system.
\newblock In \emph{Proceedings of the 15th Conference of the {E}uropean Chapter
  of the Association for Computational Linguistics: Volume 1, Long Papers},
  pages 438--449.

\bibitem[{Wixted et~al.(2018)Wixted, Goldinger, Squire, Kuhn, Papesh, Smith,
  Treiman, and Steinmetz}]{Sparse}
John~T. Wixted, Stephen~D. Goldinger, Larry~R. Squire, Joel~R. Kuhn, Megan~H.
  Papesh, Kris~A. Smith, David~M. Treiman, and Peter~N. Steinmetz. 2018.
\newblock \href {https://doi.org/10.1073/pnas.1716443115} {Coding of episodic
  memory in the human hippocampus}.
\newblock \emph{Proceedings of the National Academy of Sciences}.

\end{thebibliography}


\newpage 

\onecolumn
\appendix
\section{Appendix}

We show some responses generated by our model and grounded to a given KB. 

\bigskip 
\begin{adjustbox}{width=480pt}
\centering
\begin{tabular}{|c|c|c|c|c|}
\hline \textbf{Distance} & \textbf{Traffic Info} & \textbf{Category}  & \textbf{POI} & \textbf{Address} \\ \hline
5 miles & no traffic & coffee & Coupa & 394 Van Ness Ave  \\
5 miles & no traffic & shopping center & Midtown Shopping Center & 338 Alester Ave \\
5 miles & moderate traffic & hospital & Stanford Express Care & 214 El Camino Real \\
6 miles & moderate traffic & chinese restaurant & P.F. Changs & 669 El Camino Real  \\
2 miles & no traffic & friends house & Toms house & 580 Van Ness Ave  \\
2 miles & road block nearby & parking garage & Alto Garage R & 481 Amaranta Ave \\
2 miles & heavy traffic & chinese restaurant & Panda Express & 3842 Arrowhead Way  \\
4 miles & car collision nearby & coffee or tea place & Philz & 583 Alester Ave  \\
\hline
\end{tabular}
\end{adjustbox}

\bigskip 
\begin{tabular}{ll}
\textbf{DRIVER:}  & Find the address to a hospital or clinic \\
\textbf{CAR:}  & the closest hospital is  Stanford Express Care , located 5 miles away at 214 El Camino Real  \\
\textbf{DRIVER:}  & I would like to avoid any heavy traffic \\
\textbf{CAR:}  & there will be moderate traffic , I am sorry \\
\textbf{DRIVER:}  & Thank you  \\
\textbf{CAR:}  & great , glad I could help . have a great day  \\
\end{tabular}

\bigskip 
\begin{adjustbox}{width=250pt}
\centering
\begin{tabular}{|c|c|c|c|c|}
\hline \textbf{Time} & \textbf{Date} & \textbf{Party}  & \textbf{Event} \\ \hline
5pm& the 4th & sister & tennis activity \\
7pm& sunday & aunt & dinner \\
2pm& thursday & sister & optometrist appointment \\
11am& the 10th & Martha & yoga activity \\
3pm& the 5th & Tom & doctor appointment \\
10am& the 4th & sister & swimming activity \\
\hline
\end{tabular}
\end{adjustbox}

\bigskip 
\begin{tabular}{ll}
\textbf{DRIVER:}  & Car, find the date and time for my Tennis activity \\
\textbf{CAR:}  & your tennis activity is on the 4th and your sister is attending   \\
\textbf{DRIVER:}  & Great. Can you remind me what time the activity is ? \\
\textbf{CAR:}  & it is at 5pm \\
\textbf{DRIVER:}  & Thanks  \\
\textbf{CAR:}  & you are welcome , i am always happy to help \\
\end{tabular}

\bigskip 
\begin{adjustbox}{width=480pt}
\centering
\begin{tabular}{|c|c|c|c|c|}
\hline \textbf{Distance} & \textbf{Traffic Info} & \textbf{Category}  & \textbf{POI} & \textbf{Address} \\ \hline
2 miles & road block nearby & pizza restaurant & Dominos & 776 Arastradero Rd  \\
2 miles & heavy traffic& home& home & 10 Ames street \\
3 miles & road block nearby & coffee or tea place & Starbucks & 792 Bedoin Street \\
1 miles & no traffic & grocery store & Sigona Farmers Market & 638 Amherst St \\
4 miles & car collision nearby & parking garage & Dish Parking & 550 Alester Ave \\
4 miles & heavy traffic & hospital & Stanford Childrens Health & 899 Ames Ct \\
5 miles & no traffic & friends house & Jills house & 347 Alta Mesa Ave  \\
3 miles & car collision nearby & chinese restaurant & Tai Pan & 830 Almanor Ln \\
\hline
\end{tabular}
\end{adjustbox}

\bigskip 
\begin{tabular}{ll}
\textbf{DRIVER:}  & find starbucks \\
\textbf{CAR:}  & the nearest Starbucks is at 792 Bedoin Street and there is roads block nearby .  \\

& I sent you another route on your screen \\

\textbf{DRIVER:}  & Is it the quickest one available? I really need my morning coffee! \\
\textbf{CAR:}  & yes it is the fastest route \\
\textbf{DRIVER:}  & Thank you  \\
\textbf{CAR:}  & have a great day  ! \\
\end{tabular}

\end{document}